%% file: spdl.tex
\documentclass[12pt,fleqn]{article}

\usepackage[frozencache=true,cachedir=minted-cache]{minted} 
\usepackage{arxiv}
\usepackage[utf8]{inputenc} 
\usepackage[T1]{fontenc}    
\usepackage{url}            
\usepackage{booktabs}       
\usepackage{amsfonts}       
\usepackage{nicefrac}       
\usepackage{microtype}      

\usepackage{graphicx}
\usepackage[numbers]{natbib}
\usepackage{doi}
\usepackage{pgfplots}
\usepackage{listings}
\usepackage{algorithm}
\usepackage{algorithmicx,algpseudocode}
\usepackage[labelfont=bf, figurename=Fig.]{caption}
\usepackage{subcaption}
\usepackage{xcolor}
\usepackage{amsmath}
\usepackage{mathtools}
\usepackage{tabularx}
\usepackage{hyperref}       

\lstdefinestyle{CStyle}{
    backgroundcolor=\color{backgroundColour},   
    commentstyle=\color{mGreen},
    keywordstyle=\color{magenta},
    numberstyle=\tiny\color{mGray},
    stringstyle=\color{mPurple},
    basicstyle=\footnotesize,
    breakatwhitespace=false,         
    breaklines=true,                 
    captionpos=b,                    
    keepspaces=true,                 
    numbers=left,                    
    numbersep=5pt,                  
    showspaces=false,                
    showstringspaces=false,
    showtabs=false,                  
    tabsize=2,
    language=C
}
\title{Performance Optimization of Deep Learning Sparse Matrix Kernels on Intel Max Series GPU}

\author{ \href{https://orcid.org/0000-0002-5449-1779}{\includegraphics[scale=0.06]{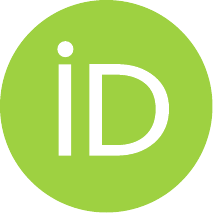}\hspace{1mm}Mohammad Zubair}
        \\
	Old Dominion University\\
	Norfolk, Virginia, USA\\
	\texttt{zubair@cs.odu.edu} \\
	\And
	\href{https://orcid.org/0009-0007-0096-2321}{\includegraphics[scale=0.06]{orcid.pdf}\hspace{1mm}Christoph Bauinger} \\
	Intel Corporation \\
	Santa Clara, CA, USA \\
	\texttt{christoph.bauinger@intel.com} \\
}

\renewcommand{\shorttitle}{}

\begin{document}
\maketitle

\begin{abstract}
In this paper, we focus on three sparse matrix operations that are relevant for machine learning applications, namely, the sparse-dense matrix multiplication (SPMM), the sampled dense-dense matrix multiplication (SDDMM), and the composition of the SDDMM with SPMM, also termed as FusedMM.  
We develop optimized implementations for SPMM, SDDMM, and FusedMM operations utilizing Intel oneAPI's Explicit SIMD (ESIMD) SYCL extension API. 
In contrast to CUDA or SYCL, the ESIMD API enables the writing of explicitly vectorized kernel code. Sparse matrix algorithms implemented with the ESIMD API achieved performance close to the peak of the targeted Intel Data Center GPU. We compare our performance results to Intel's oneMKL library on Intel GPUs and to a recent CUDA implementation for the sparse matrix operations on NVIDIA's V100 GPU and demonstrate that our implementations for sparse matrix operations outperform either.

\end{abstract}


\section{Introduction}\label{sec:introduction}
The key to building and utilizing increasingly large machine learning models is efficient implementations of the training and inferencing on emerging high-performance computing architectures, mainly graphics processing units (GPUs). Due to their high throughput in single precision (FP32) and reduced precision arithmetic (FP8, FP16, BF16, TF32, etc.), which is typically delivered by specialized hardware built into the devices, GPUs have become the standard workhorse for machine learning applications~\cite{pytorchcuda,tensorflowintel}. 
In this context, three sparse matrix operations are of particular interest. Namely, the sparse-dense matrix multiplication (SPMM), the sampled dense-dense matrix multiplication (SDDMM), and the composition of the SDDMM with SPMM, also termed as FusedMM operation~\cite{Trevor2019,Trevor2020,Trevor2022megablocks,adaptive2019,FusedMM2021,dist2022}. The SPMM operation involves the multiplication of a sparse matrix with a dense matrix, resulting in a dense matrix. The SDDMM, in turn, multiplies a dense matrix with another dense matrix with the constraint that not all the entries of the resultant matrix are needed. In other words, the output matrix of the SDDMM operation is sparse. The sparsity structure is given as input to the SDDMM operation that specifies what entries of the output matrix need to be computed. The FusedMM operation merges the SDDMM and SPMM operations into a single operation, where the output of the SDDMM operation, which is a sparse matrix, is used as input to the following SPMM operation. Fusing the operations can increase the performance and can thus be beneficial for applications such as sparse transformer~\cite{Trevor2020}, where SPMM follows the SDDMM operation.

Sparse matrix operations have been extensively studied in scientific computing~\cite{saad_book}. Methods such as finite element (FE), finite volume (FV), or finite difference (FD) results in sparse matrices with a high degree of sparsity (i.e., the fraction of zeros in the matrix). More precisely, in these methods, the density of the matrices, which describes the fraction of non-zero entries in a matrix, is often less than $1\%$ of the overall number of entries. In scientific applications, sparse matrices are typically involved in solving linear sparse systems that require either sparse matrix-vector operations or sparse matrix factorizations~\cite{saad_book,DufER86,GeoL81a}. On the other hand, the sparse matrices arising in machine learning are relatively dense, where the density can vary from $10\%$ to $30\%$~\cite{Trevor2019,Trevor2020}.
Nevertheless, the matrices are sufficiently sparse --- especially in the cases with a density close to $10\%$ --- to warrant exploration of sparse matrix operations with compressed storage formats~\cite{Trevor2020,adaptive2019}. 

Several researchers have explored efficient implementations of sparse matrix operations, including SDDMM, SPMM, and FusedMM, on emerging high-performance architectures GPUs~\cite{Trevor2020,adaptive2019,Trevor2022megablocks,Fused2020}. The major challenge in optimizing the performance of sparse matrix operations on GPUs is effectively utilizing the available memory bandwidth. We found that an implicit SIMD programming environment such as CUDA~\cite{nvidia-cudapg} or SYCL~\cite{sycl2020}, where the compiler performs the vectorization implicitly, introduces difficulties when trying to achieve performance close to the theoretical peak hardware capability. Thus, we optimize sparse matrix operations utilizing Intel oneAPI's Explicit SIMD SYCL extension (ESIMD) API~\cite{esimdapi}.
In contrast to CUDA or SYCL, the ESIMD API enables the writing of explicitly vectorized kernel code. Due to this, it allows more precise control over register usage and better handles thread divergence compared to CUDA or SYCL. In prior contributions~\cite{zubair2023}, it was shown that kernel code written with the ESIMD API can perform close to the peak of the targeted Intel Data Center GPU hardware~\cite{maxgpu}. The main disadvantage of ESIMD, in contrast to SYCL, is the lack of support for non-Intel GPUs.  

In this paper, we develop optimized implementations for the SPMM, SDDMM, and FusedMM operations for Intel Data Center GPUs utilizing the ESIMD API. In Section~\ref{sec:bg}, we describe the sparse matrix operations, Intel Data Center GPU, and Intel ESIMD API. The GPU algorithms for the sparse matrix operations are covered in Section~\ref{sec:approach}.
Section~\ref{sec:implementation} then demonstrates the optimized ESIMD implementations of the matrix operations. Finally, Section~\ref{sec:experiments} compares our results with a recent implementation for the two sparse matrix operations on an NVIDIA V100 GPU presented in~\cite{Trevor2020} and shows that we outperform the reference implementation~\cite{Trevor2020} on V100 by up to a factor 10. In addition, with our implementation, we increase the device utilization from up to 27\% of the theoretical single-precision peak of a V100 GPU (cf.~\cite{Trevor2020}) to up to 49\% on the Intel hardware. Finally, we compare our results to the implementations of the sparse operations available in Intel's oneMKL library and show that our performance outperforms the state-of-the-art MKL implementations by up to a factor of 3.4 on Intel's Data Center GPU.

\section{Background} \label{sec:bg}
In this section, we give an overview of the relevant sparse matrix operations, introduce the ESIMD API as well as the targeted Intel Data Center GPU Max 1550.
\subsection{Sparse Matrix Operations}
Table~\ref{table:notation} shows the notation used in this paper to describe the sparse matrix operations. It also lists the matrix sizes investigated in the reference implementation in~\cite{Trevor2020} and which commonly occur in deep learning networks.
\begin{table}
\centering
\begin{tabularx}{\textwidth}{l|X}
$M \in \mathbb{N}$& Value of $M$ ranges from $1024 \eqqcolon 1k$ to $32768\eqqcolon32k$ \\
$K \in \mathbb{N}$& Value of $K$ ranges from $1024\eqqcolon1k$ to $8192 \eqqcolon 8k$ \\
$N \in \mathbb{N}$& Value of $N$ is typically $32$ or $128$ \\
$\mathbf{A}\in \mathbb{R}^{M \times K}$ & A sparse matrix of size $M \times K$ \\
$\mathbf{B}\in \mathbb{R}^{K \times N}$ & A dense matrix of size  $K \times N$ \\
$\mathbf{B}^T \in \mathbb{R}^{N \times K}$ & The transpose of $\mathbf{B}$, with size $N \times K$ \\ 
$\mathbf{D}\in \mathbb{R}^{K \times N}$ & A dense matrix of size  $K \times N$ \\
$\mathbf{C}\in \mathbb{R}^{M \times N}$ & A dense matrix of size  $M \times N$ \\
$\mathbf{E}\in \mathbb{R}^{M \times N}$ & A dense matrix of size  $M \times N$ \\
$\mathbf{I}_A \in \mathbb{R}^{M \times K}$ & A sparse matrix of size $M \times K$, where all the non-zero entries have a value of 1 \\
$\mathbf{O}_i\in \mathbb{R}^{1 \times \kappa}$ & The row $i$, $1 \leq i \leq \mu$, of any matrix $\mathbf{O} \in \mathbb{R}^{\mu \times \kappa}$. \\
$\mathbf{O}_{i,j}\in \mathbb{R}$ & The element with the row index $i$, $1 \leq i \leq \mu$, and the column index $j$, $1 \leq j \leq \kappa$, of any matrix $\mathbf{O} \in \mathbb{R}^{\mu \times \kappa}$. \\
$\alpha \in [0,1) \subset \mathbb{R}$ & Sparsity. The fraction of matrix elements which are zero. \\
$nnz \in \mathbb{N}$ & number of non-zero elements in the matrix $\mathbf{A}$. Computed as $nnz = M K \beta$ \\
\end{tabularx}
\vspace{5pt}
\caption{Summary of the relevant quantities and the notation used in the description of the algorithms.}
\label{table:notation}
\end{table}

\subsubsection*{SDDMM}
The SDDMM operation is defined as follows.
\begin{eqnarray} \label{eq:sddmm}
\mathbf{A} & = & \mathbf{C} \mathbf{B}^T \odot \mathbf{I}_A
\end{eqnarray}
For this operation, a dense matrix $\mathbf{C}$ is multiplied with a dense matrix $\mathbf{B}^T$ followed by an element-wise product, which is indicated by the operation $\odot$, with a matrix $\mathbf{I}_A$. In a typical implementation, a dot product of a row of the matrix $\mathbf{C}$ with a column of $\mathbf{B}^T$ is computed only at a location of a non-zero entry in $\mathbf{I}_A$.

\subsubsection*{SPMM} 
The SPMM operation, on the other hand, multiplies a sparse matrix $\mathbf{A}$ with a dense matrix $\mathbf{B}$, resulting in a dense matrix $\mathbf{C}$, as shown below.
\begin{eqnarray}
\mathbf{C} & = & \mathbf{A} \mathbf{B} \label{eq:spmm}
\end{eqnarray}

\subsubsection*{FusedMM}
The FusedMM operation is a composition of SDDMM with SPMM. The composite operation avoids storing the result of the SDDMM operation explicitly and can have a better performance compared to an implementation with the two operations applied one after another. 
\begin{eqnarray}
\mathbf{E} & = & (\mathbf{C} \mathbf{B}^T \odot \mathbf{I}_A) \mathbf{D} \label{eq:fusedmm} 
\end{eqnarray}

\subsubsection*{Matrix Storage Layout}
All dense matrices are stored in row-major order. The sparse matrices are stored in a compressed sparse row format, which is described in what follows.
 
The sparsity of the matrix $\mathbf{A}$ is denoted by a value $1 > \alpha \in \mathbb{R}$, which indicates the fraction of zero-entries in the matrix $\mathbf{A}$. In deep learning networks~\cite{Trevor2019,Trevor2020}, the sparsity typically varies from $0.7$ to $0.9$. The compressed sparse row (CSR) format~\cite{saad_book} avoids the explicit storage of zeros in the matrix. In this format, only non-zero entries in the matrix are stored along with two integer arrays to resolve the row and column indices of the non-zero values. This format reduces the memory footprint and operation count for sparse matrix operations at the expense of indirect addressing. The CSR format for the matrix $\mathbf{A}$ consists of three one-dimensional arrays: $avalue \in \mathbb{R}^{nnz}$, $ia \in \mathbb{N}^{M+1}$, and $ja \in \mathbb{N}^{nnz}$. The size of the $avalue$ and $ja$ arrays is $nnz$, the number of non-zeros in the sparse matrix. The array $avalue$ contains the non-zero entries in $\mathbf{A}$ in row-major order, and the $ja$ array contains column indices of those values. The array $ia$ is of size $M+1$ whose $i$-th entry indicates the index in $avalue$ and $ja$ where the $i$-th row of $\mathbf{A}$ starts. The array $ia$ includes a fictitious
$M+1$-th entry to facilitate easy traversal of the elements through the last row $M$. 
Figure~\ref{fig:matstructure} shows a sample sparse matrix with the corresponding CSR arrays.

\begin{figure}
\centering
\begin{subfigure}[t]{0.2\textwidth}
\centering
\[
\begin{bmatrix}
~ & ~ & 1 & 2 \\ 
~ & ~ & 3 & ~ \\ 
4 & 5 & ~ & ~ \\ 
6 & ~ & ~ & ~  \\
7 & ~ & 8 & 9 
\end{bmatrix}
\]
\phantomsubcaption
\end{subfigure}
\begin{subfigure}[t]{0.2\textwidth}
\centering
\begin{eqnarray*}
ia & =  & \left[0,2,3,5,6,9 \right] \\
ja & = &  \left[2,3,2,0,1,0,0,2,3  \right] \\
avalue & = &  \left[1,2,3,4,5,6,7,8,9\right] 
\end{eqnarray*}
\phantomsubcaption
\end{subfigure}
\caption{An example of a sparse matrix of size $M \times K$ with $M = 5$ and $K=4$ on the left-hand side. Note that only non-zero entries are shown. The three CSR arrays $ia$, $ja$, and $avalue$ required for storing the sparse matrix are presented on the right-hand side. }
\label{fig:matstructure}     
\end{figure}

\subsection{Intel Data Center GPU Max 1550}\label{sec:maxgpu}
The code in the present contribution targets the nascent Intel Data Center GPU Max 1550~\cite{maxgpu}. In what follows, this GPU is briefly introduced. 
The Intel Data Center GPU Max 1550, which is shortened to {\em Intel GPU} in what follows, is a HPC accelerator with 128 gigabyte (GB) high-bandwidth memory (HBM), and a theoretical peak FP32, and FP64 throughput of approximately 54 tera floating-point operations per second (Tflops/s) which is delivered by 1024 so-called vector engines. For machine learning applications the device includes so-called matrix engines (XMX) delivering a theoretical peak performance of 832 Tflops/s on the bfloat16 data type.

From a performance optimization point of view, it is important to note that the device is not one monolithic piece of hardware but consists of two {\em Xe stacks}. This is relevant considering that, i) the interconnect between the stacks is slower than the access to HBM memory, and ii) cross-stack accesses are not cached. It is therefore recommended for most applications to scale explicitly to the two stacks using, e.g., MPI or multiple queues. Thus, in what follows, we focus on the performance of a single stack of the Intel GPU. For more information on the Intel hardware, we refer to~\cite{maxgpu,xearchitecture}

\subsection{Intel oneAPI/ESIMD}\label{sec:oneapi-esimd}
The ESIMD API~\cite{esimdapi,esimd2023_1} is an extension to the SYCL standard developed specifically for Intel GPUs. It is based on Intel's GPU Instruction Set Architecture (ISA). Whereas SYCL relies on the compiler for the vectorization along the work-items within a sub-group~\cite{intelthreadmapping} (cf. ``warp'' in CUDA), in ESIMD the code uses \texttt{simd} objects (see Fig.~\ref{fig:example-ESIMD}) for explicit vectorization. These \texttt{simd} objects enable the vectorization over SIMD-sizes different from the sub-group size. Thus, ESIMD offers finer control over the vectorization compared to standard SYCL. Since the \texttt{simd} objects are mapped to the registers, ESIMD permits fine control over register usage. Additionally, ESIMD provides APIs for explicit memory load, store, and prefetch operations with parameters to control the caching behavior, and it simplifies the management of divergent branches in kernel code. 

In contrast to SYCL, ESIMD does not have the concept of sub-groups. Since each work-item in ESIMD utilizes explicit vectorization, the notion of sub-groups is not required and each work-item in ESIMD represents, in some sense, a SYCL sub-group with variable size. Thus, while a work-item in SYCL is equivalent to a thread in CUDA, a work-item in ESIMD differs.

As shown in Fig.~\ref{fig:example-ESIMD}, when using Intel oneAPI DPC++, an ESIMD kernel is launched with the "SYCL\_ESIMD\_KERNEL" property.

\begin{figure}
\begin{minted}[frame=lines, linenos, fontsize=\footnotesize]
{c++}
#define SIMD_LEN 16

void vecAdd(nd_item<1> item, double *a, double *b, double *c) {
    const int i = item.get_global_id(0)*SIMD_LEN;
    simd<double, SIMD_LEN> va, vb, vc;
    va.copy_from(a + i);
    vb.copy_from(b + i);
    vc = va + vb;
    vc.copy_to(c + i);
}

int main() {
    queue Q(gpu_selector_v);
    double *d_a = malloc_device<double>(16*32, Q);
    double *d_b = malloc_device<double>(16*32, Q);
    double *d_c = malloc_device<double>(16*32, Q);
    Q.parallel_for(nd_range<1>(16*32/SIMD_LEN, 32), 
    [=](nd_item<1> item) SYCL_ESIMD_KERNEL {
        vecAdd(item, d_a, d_b, d_c);
    }); 
}
\end{minted}
    \caption{An example of a vector addition written in ESIMD.}
    \label{fig:example-ESIMD}
\end{figure}

\section{Overview of the Algorithms}\label{sec:approach} 
In this section, we discuss our approach to parallelizing sparse matrix operations without giving details on how these algorithms are realized on a specific architecture. The implementation details specific to a GPU architecture are covered in the next section, where we account for the underlying vector architecture, number of registers, cache behavior, etc.

\subsection{Parallelizing SDDMM}
\begin{figure}
    \centering
    \includegraphics[scale=0.6]{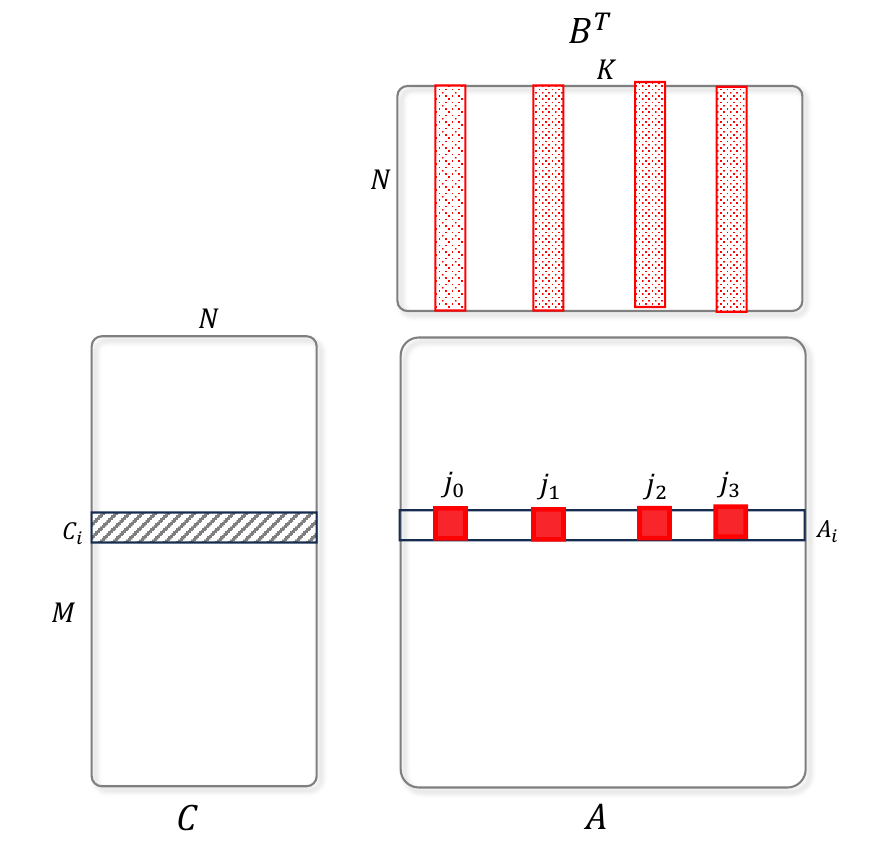}
    \caption{Illustration of SDDMM implementation for computing non-zeros of $\mathbf{A}_i$. A single thread $i$ computes all non-zeros of $\mathbf{A}_i$.} 
    \label{fig:sddmm_simple}
\end{figure}

The SDDMM operation multiplies two dense matrices, $\mathbf{C}$ and $\mathbf{B}^T$, and stores the result in a sparse matrix $\mathbf{A}$. The parallelization is performed along the elements of the resulting matrix $\mathbf{A}$. 
A total of $M$ threads are assigned to work concurrently, where a thread $i$, $1 \leq i \leq M$, computes all non-zeros of the row $\mathbf{A}_{i}$. Figure~\ref{fig:sddmm_simple} illustrates this computation for $\mathbf{A}_i$ that has four non-zeros at column indices $j_0, j_1, j_2$, and $j_3$. The non-zero at $j_0$ is computed by multiplying the row $\mathbf{C}_i$ with a column $j_0$ of $\mathbf{B}^T$. The rest of the non-zeros are computed similarly. A high-level description of the algorithm for computing $\mathbf{A}_{i}$ is outlined in Algorithm~\ref{alg:sddmm1}. In contrast to the above description, where we implicitly assumed a regular matrix layout for $\mathbf{A}$, we use the CSR format for $\mathbf{A}$ in the high-level description of the algorithm. Please note that input to the SDDMM algorithm is $\mathbf{B}$. However, we are interested in the product of $\mathbf{C}$ with $\mathbf{B}^T$ (cf. Equation~\eqref{eq:sddmm}).

\begin{algorithm}[ht]
\small
\caption{\textsc{SDDMM-ROW}($ia, ja, avalues, B, C, N, \texttt{i}$)}
\label{algo:sddmm}
\begin{algorithmic}[1]
\State {Initialize: $\mathbf{C}[i,:] \leftarrow 0$ }
\State {$nnzr \leftarrow ia[i+1]-ia[i]$}
\For {$ j \leftarrow 0$ \textbf{to} $nnzr-1$}
\State {$k \leftarrow  \texttt{ja}[ia[i]+j]$}
\State {$dp \leftarrow  0$}
\For {$ l \leftarrow 0$ \textbf{to} $N-1$}
\State {$dp \leftarrow  dp + \mathbf{C}[i,l] * \mathbf{B}[k,l]$}
\EndFor
\State {$\texttt{avalues}[ia[i]+j]  \leftarrow  dp $}
\EndFor
\State \Return $\texttt{avalues}$
\end{algorithmic}
\label{alg:sddmm1}
\end{algorithm}

\begin{figure}
    \centering
    \includegraphics[scale=0.6]{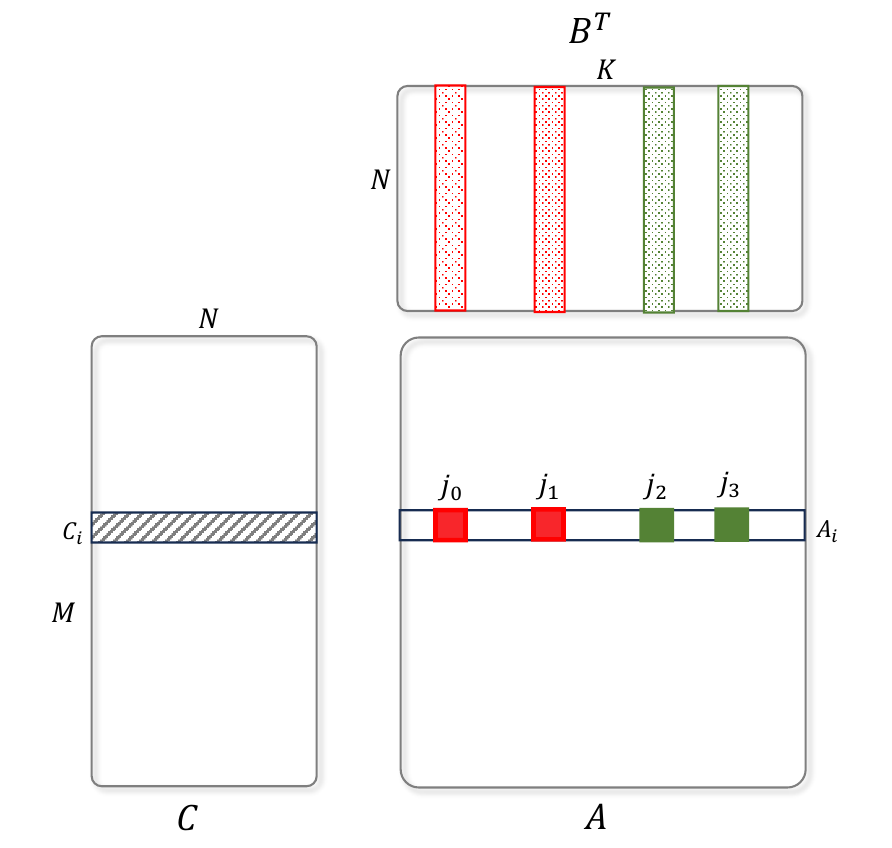}
    \caption{SDDMM implementation with two threads assigned to compute a of $\mathbf{A}$. }
    \label{fig:sddmm_simple2}
\end{figure}

The parallelism in SDDMM operation can be increased by assigning more than one thread to compute non-zeros of $\mathbf{A}_i$. For example, we can assign two threads to compute the non-zeros of a row $\mathbf{A}_i$. In this case, we need a total of $2M$ threads where two threads $2i$ and $2i+1$ collectively compute all non-zeros of $\mathbf{A}_i$, with each thread computing half of the non-zeros. Figure~\ref{fig:sddmm_simple2} illustrates this computation for $\mathbf{A}_i$ that has four non-zeros at column indices $j_0, j_1, j_2$, and $j_3$. Thread $2i$ computes non-zeros at column indices $j_0$ and $j_1$; and thread $2i+1$ computes non-zeros at column indices $j_2$ and $j_3$. The SDDMM Algorithm~\ref{alg:sddmm1} for one thread per row of the output matrix can be easily adjusted to work with two threads per row. 

\subsection{Parallelizing SPMM}
\begin{figure}[ht]
 \centering
\includegraphics[scale=0.6]{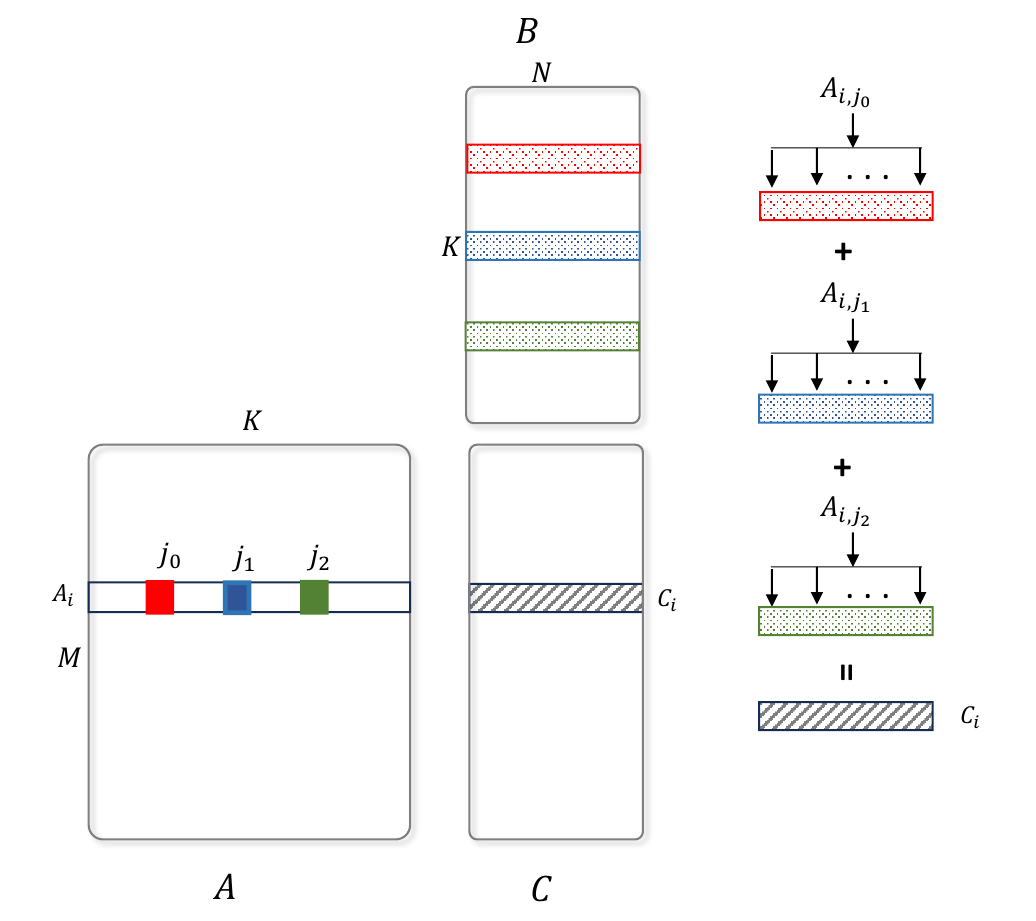}
 \caption{A simple example to illustrate multiplication of the row $\mathbf{A}_{i}$ with $\mathbf{B}$ to compute the row $\mathbf{C}_{i}$ in the SPMM algorithm.} 
\label{fig:spmm-alg}
\end{figure}

The SPMM operation multiplies a sparse matrix $\mathbf{A}$ with a dense matrix $\mathbf{B}$ to compute a dense matrix $\mathbf{C}$. The output matrix $\mathbf{C}$ is of size $M \times N$. We parallelize the computation over the rows of the matrix $\mathbf{C}$. A total of $M$ threads are assigned to work concurrently, where a thread $i$, $1 \leq i \leq M$, computes row $\mathbf{C}_{i}$. This computation requires multiplying a sparse row $\mathbf{A}_{i}$ with selected elements of the $\mathbf{B}$ matrix to compute $\mathbf{C}_{i}$. We assume there are three non-zeros in row $\mathbf{A}_{i}$ at column indices $j_0$, $j_1$ and $j_2$, as shown in Figure~\ref{fig:spmm-alg}. It implies that we need rows $\mathbf{B}_{j_0}$, $\mathbf{B}_{j_1}$, and $\mathbf{B}_{j_2}$ to compute the row $\mathbf{C}_{i}$. We implement this computation by multiplying $\mathbf{A}_{i,J_0}$ with all elements in the row $\mathbf{B}_{0}$. The result is added to the multiplication of $\mathbf{A}_{i,j_1}$ with $\mathbf{B}_{j_1}$, and so on.

 A high-level description of the algorithm for computing a row $\mathbf{C}_{i}$ is outlined in Algorithm~\ref{fig:spmm-alg}. As before, we use the CSR format for $\mathbf{A}$ to describe the algorithm. The main loop is over the non-zero elements in $\mathbf{A}_i$. In line~4, we obtain the $j$-th non-zero value in row $i$ of $\mathbf{A}$. The column index, $k$, associated with the $j$-th non-zero value is fetched in line~5. The for loop starting at line~6 performs a SAXPY computation~\cite{laug}, which is simply multiplying a scalar ($s$) with a vector ($\mathbf{B}_k$) and adding the result to another vector ($\mathbf{C}_i$). 

\begin{algorithm}[ht]
\small
\caption{\textsc{SPMM-ROW}($ia, ja, avalues, B, C, N, \texttt{i}$)}
\label{algo:spmm}
\begin{algorithmic}[1]
\State {Initialize: $C[i,:] \leftarrow 0$ }
\State {$nnzr \leftarrow ia[i+1]-ia[i]$}
\For {$ j \leftarrow 0$ \textbf{to} $nnzr-1$}
\State {$s \leftarrow  \texttt{avalues}[ia[i]+j]$}
\State {$k \leftarrow  \texttt{ja}[ia[i]+j]$}
\For {$ l \leftarrow 0$ \textbf{to} $N-1$}
\State {$C[i,l] \leftarrow  C[i,l] + s * B[k,l]$}
\EndFor
\EndFor
\State \Return $C$
\end{algorithmic}
\end{algorithm}

\subsection{Parallelizing FusedMM}
\begin{figure}[t]
 \centering
\includegraphics[scale=0.6]{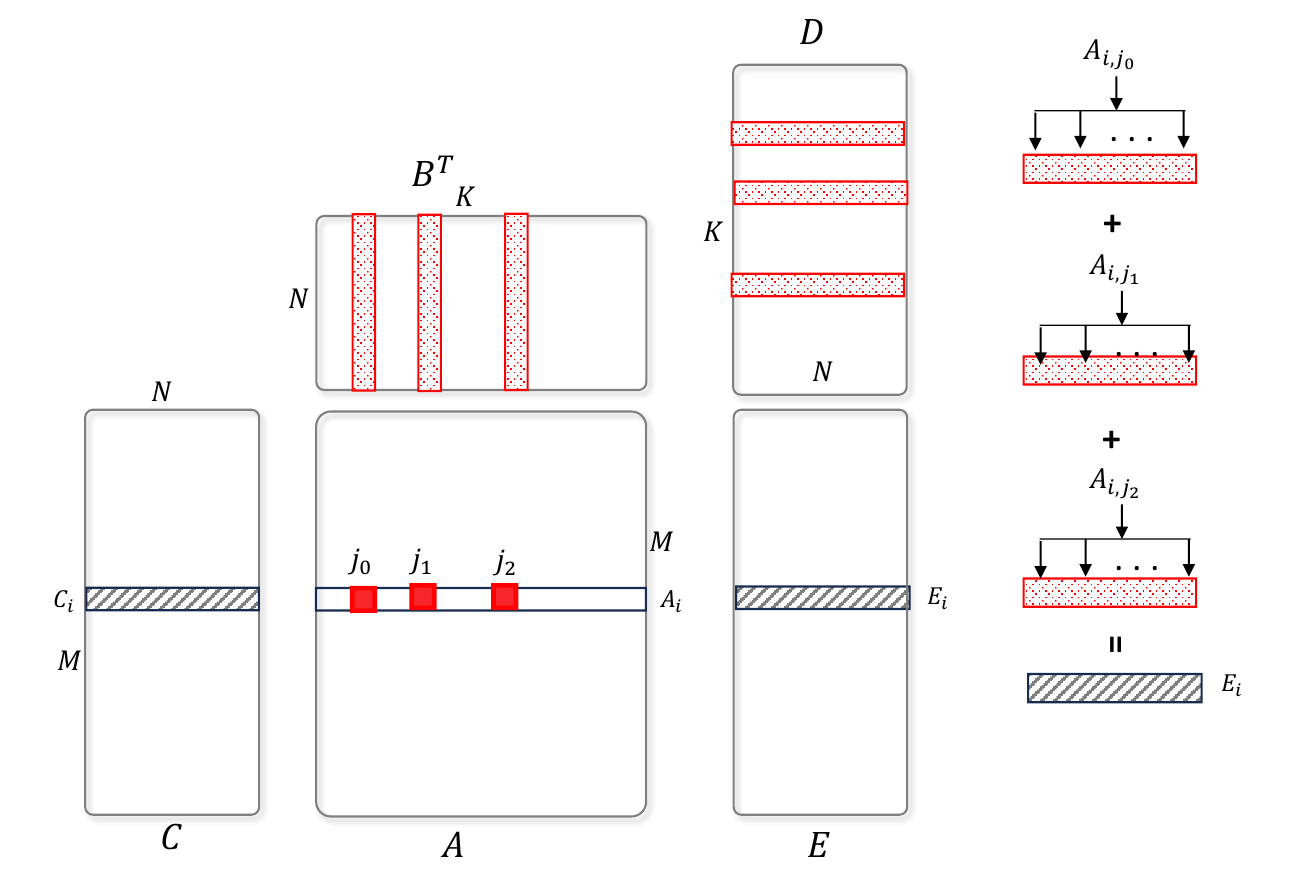}
 \caption{A simple example to illustrate FusedMM operation.} 
\label{fig:fusedMM-alg}
\end{figure}

The parallel fused version of the SDDMM and SPMM operation is straightforward. We consider a single thread per rows version of the SDDMM algorithm, see Figure~\ref{fig:sddmm_simple}. Each thread is responsible for computing non-zero values for a row of sparse matrix $\mathbf{A}$. Once a thread has computed all the non-zero values of a row (note we skip storing these values in the global array $\mathbf{A}$), we start the SPMM operation as outlined earlier; see Figure~\ref{fig:spmm-alg}. Figure~\ref{fig:fusedMM-alg} illustrates the FusedMM operation.   A high-level description of the algorithm for computing a row $\mathbf{E}_{i}$ is outlined in Algorithm~\ref{alg:fused}, which is essentially derived from Algorithm~\ref{algo:sddmm} and Algorithm~\ref{algo:spmm}.

\begin{algorithm}[ht]
\small
\caption{\textsc{FusedMM-ROW}($ia, ja, C, B, D, E, N, \texttt{i}$)}
\label{algo:fused}
\begin{algorithmic}[1]
\State {Initialize: $C[i,:] \leftarrow 0$ }
\State {$nnzr \leftarrow ia[i+1]-ia[i]$}
\State {$arow[0:nnzr] \leftarrow 0$}
\For {$ j \leftarrow 0$ \textbf{to} $nnzr-1$}
\State {$k \leftarrow  \texttt{ja}[ia[i]+j]$}
\For {$ l \leftarrow 0$ \textbf{to} $N-1$}
\State {$arow[j] \leftarrow  arow[j] + C[i,l] * B[k,l]$}
\EndFor
\EndFor
\State {Initialize: $E[i,:] \leftarrow 0$ }
\For {$ j \leftarrow 0$ \textbf{to} $nnzr-1$}
\State {$s \leftarrow  \texttt{arow}[j]$}
\State {$k \leftarrow  \texttt{ja}[ia[i]+j]$}
\For {$ l \leftarrow 0$ \textbf{to} $N-1$}
\State {$E[i,l] \leftarrow  E[i,l] + s * B[k,l]$}
\EndFor
\EndFor
\State \Return $E$
\end{algorithmic}
\label{alg:fused}
\end{algorithm}

\section{ESIMD Implementation}\label{sec:implementation}
\subsection{Sparse Matrix Collection}
The range of matrix sizes in our collection influences the implementation. Specifically, our implementation is only guaranteed to work for $N = 32$ or $N = 128$, which is typically the batch size used in the machine learning training dataset. For performance comparison, we use the same matrix dataset that is used in~\cite{Trevor2020}. It consists of 72 tests of varying sizes (cf. Figs.~\ref{fig:sddmm_benchmark32}-~\ref{fig:fused_benchmark128}) and randomized input matrices.

\subsection{ESIMD Implementation of SDDMM}
The ESIMD implementation of the SDDMM operation is based on the algorithm covered in Section~3.1 (Algorithm~\ref{algo:sddmm}). 
In the general version of this algorithm, one or more threads are assigned to compute a row of the output sparse matrix $\mathbf{A}$. The SDDMM ESIMD kernel is launched with $M \times NT$ threads~\footnote{In our discussion, we use thread and ESIMD work item interchangeably.} with a work-group of size $NT$. A work-group $i$ consisting of $NT$ threads computes a row $i$ of the sparse matrix of $nnzr$ non-zeros. Divide the number of non-zeros to be computed, $nnzr$, for a row equally amongst $NT$ threads, where a thread works on $nnzt = \lceil nnzr/NT \rceil$ non-zeros. The number of non-zeros assigned to a thread, $nnzt$, are further partitioned into chunks, where the number of chunks is given by  $nchunks = \lceil nnzt/VLC \rceil$. Here, $VLC$ is a tuning parameter that varies in the size range of \texttt{simd} objects supported by ESIMD API. 

The SDDMM kernel code segment is shown in Figure~\ref{fig:sddmm-code-1}.
In line~3, we assign the work-group id to the variable $i$, and $j$ is assigned the local thread id that ranges from $0$ to $NT-1$. In line~5, a work-group $i$ load two consecutive values $ia[i]$ and $ia[i+1]$ into a vector of size $2$. In line~10, we load row $\mathbf{C}_i$ into  a \texttt{simd} object $regL$ of size $N$. Next, we setup a nested loop, where the outer loop is over $nchunks$ and the inside loop is over $VLC$.  In the outer loop we load column indices for the current chunk into a \texttt{simd} object, $ja\_block$, of size $VLC$. In the inside loop, we first load the column of $B$ that is needed for the current iteration (line~19-20) into $regR$. Next, we multiply the two \texttt{simd} objects, $regL$ and $regR$ and store the result in a \texttt{simd} object $res$. In line~22, we do the reduction operation on the \texttt{simd} object $res$, and store the result in position $l0$ of a \texttt{simd} object $dp$ of size $VLC$. At the end of the inside loop, we have the $VLC$ non-zero values that are stored in the appropriate location in $avalues$. The is repeated over $nchunks$, and at the end of the outer loop we have computed all the non-zero values in row $\mathbf{A}_i$ of the sparse matrix. To keep our code segment simple, we avoid the details for handling the cases when $nnzr$ is not a multiple of $NT$, and $nnzt$ is not a multiple of $VLC$.  

To hide memory latency and effectively use the vector engine, we add prefetches and unroll the inside loop as illustrated in the code segment in Figure~\ref{fig:sddmm-code-2}. The prefetch is added to avoid stalls in loading the $ja\_block$. In the outer loop iteration $l$, we prefetch the memory location from where we will load $ja\_block$ in $(l+1)$-th iteration (see line~19-20 of Figure~\ref{fig:sddmm-code-2}). We add pragma to unroll the inside loop (see line~10). 

Additionally, we template the kernel on $NT$ (not shown in the code listing). We implemented a scheme that chooses the value of $NT$ in dependence of the matrix size such that the occupancy on the device is maximized. To achieve 100\% occupancy on a single-stack of the Intel GPU, one needs to launch 4096 esimd work-items or a multiple thereof. We choose $NT$ as the minimum power of two, not larger than 16, to maximize the occupancy, which is given as follows
\begin{equation*}
    \text{Occupancy} = \frac{\frac{M\times NT}{4096}}{\lceil \frac{M\times NT}{4096} \rceil}.
\end{equation*}
The ceiled value in the denominator ensures that cases where $M\times NT > 4096$ are handled appropriately.
For example, let $M=1024$, then $NT=4$ is chosen. For $M=3072$ the choice of $NT=4$ leads to 100\% occupancy.

\begin{figure}
    \centering
\begin{minted}[frame=lines, linenos, fontsize=\footnotesize]
{c++}
#define VLC 32

const int i = item.get_group(0);
const int j = item.get_local_id(0);
simd<int, 2> lrowptr2 = lsc_block_load<int, 2>(ia + i);
const int nnzr = lrowptr2[1] - lrowptr2[0];
const int nnzt = (nnzr + NT - 1) / NT;
const int nchunks = (nnzt + VLC - 1) / VLC;

simd<float, N> regL = my_lsc_block_load<float, N>(C + i * N);
simd<int, VLC> ja_block;
simd<float, VLC> a_row;
for (int l = 0; l < nchunks; l++)
{
	int idxb = lrowptr2[0] + j * nnzt + l * VLC;
	ja_block = lsc_block_load<int, VLC>(ja + idxb);
	for (int l0 = 0; l0 < VLC; l0++ )
	{
		simd<float, N> regR =
		  my_lsc_block_load<float, N>(B + ja_block[l0] * N);
		simd<float, N> res = regL * regR;
		a_row[l0] = reduce<float, float, N>(res, std::plus<>());
	}
	lsc_block_store<float, VLC>(avalues + idxb, a_row);
}

\end{minted}
\caption{ESIMD implementation of the SDDMM operation.} 
\label{fig:sddmm-code-1}
\end{figure}

\begin{figure}
    \centering
\begin{minted}[frame=lines, linenos, fontsize=\footnotesize]
{c++}

simd<float, N> regL = my_lsc_block_load<float, N>(C + i * N);
simd<int, VLC> ja_block;
simd<float, VLC> a_row;
int idxb = lrowptr2[0] + j * nnzt ;
lsc_prefetch<int, VLC, DSZ, L1_C, L3_C>(&ja[idxb]);
for (int l = 0; l < nchunks; l++)
{
	ja_block = lsc_block_load<int, VLC>(ja + idxb);
	#pragma unroll 
	for (int l0 = 0; l0 < VLC; l0++ )
	{
		simd<float, N> regR =
		  my_lsc_block_load<float, N>(B + ja_block[l0] * N);
		simd<float, N> res = regL * regR;
		a_row[l0] = reduce<float, float, N>(res, std::plus<>());
	}
	lsc_block_store<float, VLC>(avalues + idxb, a_row);
        idxb = lrowptr2[0] + j * nnzt + (l+1) * VLC;
        lsc_prefetch<int, VLC, DSZ, L1_C, L3_C>(&ja[idxb]);	
}
\end{minted}
\caption{SDDMM kernel with prefetching and unrolling of the inside loop.} 
\label{fig:sddmm-code-2}
\end{figure}

\subsection{Implementation of SPMM}
The ESIMD implementation of SPMM is based on the Algorithm~\ref{algo:spmm}.  
As outlined there, we parallelize over the rows of matrix $\mathbf{C}$. 
The SPMM ESIMD kernel is launched with $M$ threads with 
 work-group of size one. In Algorithm~\ref{algo:spmm}, the main loop is over $nnzr$, the number of non-zero elements in $\mathbf{A}_i$. For our implementation, we partition $nnzr$ into $nchunk$ chunks, where $nchunks = \lceil nnzt/VLC \rceil$. Here, $VLC$ is a tuning parameter that varies in the size range of \texttt{simd} objects supported by ESIMD API. 
 
 The SPMM kernel code segment is shown in Figure~\ref{fig:spmm-code-1}.
 We setup a nested loop, where the outer loop is over $nchunks$ and the inside loop is over $VLC$.  In the outer loop we load column indices for the current chunk into a \texttt{simd} object, $ja\_block$, of size $VLC$ (line~18). In the outer loop, in line~20 we load a chunk of size $VLC$ of non-zero values from the row of sparse matrix into a \texttt{simd} object $arow$.    
 In the inside loop, we load the $j0$-th element from $arow$, a scalar value into $s$. In line~25, we load the required row of $\mathbf{B}$ of size $N$ into a \texttt{simd} object $b\_row$. Next, we multiply the scalar value $s$ with  $b\_row$ and accumulate the result in a \texttt{simd} object $c\_row$ of size $N$. At the end of the outer loop, $c\_row$ holds the values of $\mathbf{C}_i$, which is then stored in $\mathbf{C}$ in line~30.   To keep our code segment simple, we avoid the details for handling the case when $nnzr$ is not a multiple of $VLC$. 
 We add prefetching and unrolling of the inside loop in the SPMM code similar to SDDMM code, see Figure~\ref{fig:spmm-code-2}.
 For improving occupancy, particularly for small-size matrices, having more threads in the work-group is desirable. The code in Figure~\ref{fig:spmm-code-2} can be easily adjusted to support $NT$ threads in a work-group, which will compute $NT$ rows of the output matrix $\mathbf{C}$. Note that the kernel will still be launched with a total of $M$ threads.

\begin{figure}
    \centering
\begin{minted}[frame=lines, linenos, fontsize=\footnotesize]
{c++}
#define VLC 32

const int i = item.get_group(0);
simd<int, 2> lrowptr2 = lsc_block_load<int, 2>(ia + i);
const int nnzr = lrowptr2[1] - lrowptr2[0];
const int nchunks = (nnzr + VLC - 1) / VLC;

simd<int, VLC> ja_block;
simd<float, VLC> a_row;
simd<float, N> c_row;
simd<float, N> b_row;
c_row = 0.0;

for (int l = 0; l < nchunks; l++)
{
    int idxb = lrowptr2[0] + l * VLC;
    ja_block = lsc_block_load<int, VLC>(ja + idxb);
    ja_block = ja_block * N;
    a_row = lsc_block_load<float, VLC>(avalues + idxb);
    
    for (int j0 = 0; j0 < VLC; j0++)
    {
        float s  = a_row[j0];
        int colid = ja_block[j0];
        b_row.copy_from(B + colid);
        c_row = c_row + s * b_row;
    }
}
c_row.copy_to(C + i * N);

\end{minted}
\caption{SPMM Kernel.} 
\label{fig:spmm-code-1}
\end{figure}

\begin{figure}
    \centering
\begin{minted}[frame=lines, linenos, fontsize=\footnotesize]
{c++}
simd<int, VLC> ja_block;
simd<float, VLC> a_row;
simd<float, N> c_row;
simd<float, N> b_row;
c_row = 0.0;

int idxb = lrowptr2[0] ;
lsc_prefetch<int, VLC, DSZ, L1_C, L3_C>(&ja[idxb]);
lsc_prefetch<float, VLC, DSZ, L1_C, L3_C>(&avalues[idxb]);

for (int l = 0; l < nchunks; l++)
{
    ja_block = lsc_block_load<int, VLC>(ja + idxb);
    ja_block = ja_block * N;
    a_row = lsc_block_load<float, VLC>(avalues + idxb);
    idxb = lrowptr2[0] + (l + 1) * VLC;
    lsc_prefetch<int, 32, DSZ, L1_C, L3_C>(&ja[idxb]);
    lsc_prefetch<float, 32, DSZ, L1_C, L3_C>(&avalues[idxb]);
#pragma unroll
    for (int j0 = 0; j0 < VLC; j0++)
    {
        float s  = a_row[j0];
        int colid = ja_block[j0];
        b_row.copy_from(B + colid);
        c_row = c_row + s * b_row;
    }
}
c_row.copy_to(C + i * N);

\end{minted}
\caption{SPMM Kernel with prefetching and unrolling of the inside loop.} 
\label{fig:spmm-code-2}
\end{figure}

\subsection{Implementation of FusedMM}
We can easily fuse the SDDMM with SPMM for a work-group with $NT=1$ thread. The choice of $NT=1$ ensures the two operations SDDMM and SPMM require an identical number of total threads. Recall that the SDDMM kernel is launched with $M \times NT$ threads and that the SPMM kernel is launched with $M$ threads (independent of $NT$). The fused kernel for $NT=1$ is shown in Figure~\ref{fig:fused-code}. The inputs to the kernel are $\mathbf{C}$, $\mathbf{B}$, $\mathbf{D}$ matrices, along with the two arrays, $ia$ and $ja$, that capture the sparsity of the $\mathbf{A}$ matrix, see Equation~\ref{eq:fusedmm}. The output is the $\mathbf{E}$ matrix. 
One can observe that the first part of the FusedMM code~\ref{fig:fused-code}, until line~28, is a copy of the SDDMM kernel code without storing the $a\_row$ into $avalues$ (line~18 of Figure~\ref{fig:sddmm-code-2}). Once the data in $a\_row$ is ready, we start the SPMM kernel. More specifically, we took the inside loop of the SPMM kernel, line~20-26 of Figure~\ref{fig:spmm-code-2} and used it in the fused kernel at line~31-37 after taking into account that we are working with different matrices. At the end of the outer loop in~\ref{fig:fused-code} we copy $e\_row$ to $\mathbf{E}_i$ similar to line~28 of the SPMM kernel, Figure~\ref{fig:spmm-code-2}.

\begin{figure}
    \centering
\begin{minted}[frame=lines, linenos, fontsize=\footnotesize]
{c++}
const int i = item.get_group(0);
simd<int, 2> lrowptr2 = lsc_block_load<int, 2>(ia + i);
const int nnzr = lrowptr2[1] - lrowptr2[0];
const int nchunks = (nnzr + VLC - 1) / VLC;

simd<float, N> regL = my_lsc_block_load<float, N>(C + i * N);
simd<int, VLC> ja_block;

int idxb = lrowptr2[0];
lsc_prefetch<int, 32, DSZ, L1_C, L3_C>(&ja[idxb]);

simd<float, VLC> a_row;
simd<float, N> e_row;
simd<float, N> d_row;
e_row = 0.0;

for (int l = 0; l < nchunks - 1; l++)
{
    ja_block = lsc_block_load<int, VLC>(ja + idxb);
    ja_block = ja_block * N;

#pragma unroll
    for (int l0 = 0; l0 < VLC; l0++)
    {
        simd<float, N> regR = my_lsc_block_load<float, N>(B + ja_block[lo]);
        simd<float, N> res = regL * regR;
        a_row[l0] = reduce<float, float, N>(res, std::plus<>());
    }

#pragma unroll
    for (int j0 = 0; j0 < VLC; j0++)
    {
        float s = a_row[j0];
        int colid = ja_block[j0];
        d_row.copy_from(D + colid);
        e_row = e_row + s * d_row;
    }
    idxb = lrowptr2[0] + (l + 1) * VLC;
    lsc_prefetch<int, 32, DSZ, L1_C, L3_C>(&ja[idxb]);
}
e_row.copy_to(E + i*N);

\end{minted}
\caption{FusedMM Kernel with prefetching and unrolling of the inside loop.} 
\label{fig:fused-code}
\end{figure}

\section{Experiments}\label{sec:experiments}
In this section we compare our implementations to the implementations available in MKL. All tests were performed on a single stack of an Intel Data Center GPU Max 1550 on an Intel-internal test system. The code was compiled with Intel's icpx compiler (2023.2.0.20230721), which is included in Intel's oneAPI, version 2023.2.1. We used the compile options "-fsycl", "-O3" and, in the case of the oneMKL tests, "-qmkl". Further, we used an unreleased engineering GPU driver for the tests~\footnote{agama-ci-devel/682.16}. 

Figures~\ref{fig:sddmm_benchmark32} and~\ref{fig:sddmm_benchmark128} show the performance of our SDDMM implementation compared to oneMKL's (included in oneAPI version 2023.2.1~\cite{oneapi20232}) dense blas::gemm implementation~\cite{mklblasgemm}. This comparison is highly disadvantageous for oneMKL since 70\%-90\% (depending on the sparsity of the workload) of the computed flops in a dense matrix-matrix multiplication are irrelevant for the output of SDDMM. Thus, although oneMKL blas::gemm achieves a peak performance of approximately 25,000 gigaflops per second (Gflops/s) in the dense matrix multiplication, our specialized SDDMM implementation can outperform it. The reason we compare to this oneMKL function is that there is not yet a dedicated SDDMM functionality included in oneMKL. Overall, our implementation shows an average speedup of 1.84x compared to oneMKL's dense gemm and a maximum speedup of 3.4x.

Compared to the implementation shown in~\cite{Trevor2020}, we increase the peak performance from approximately 2,700 Gflops/s (achieved on the 4k/1k/128/70\% case) to close to 10,700 Gflops/s (achieved on the 12k/4k/128/70\% case). Further, the average performance gains achieved compared to~\cite{Trevor2020} in the SDDMM case is a factor of approximately 4.5x with a maximum relative performance increase of approximately 10x (achieved on the 32k/8k/32/90\% case).
\include{figs/sddmm_benchmark}

Figures~\ref{fig:spmm_benchmark32} and~\ref{fig:spmm_benchmark128} show the performance of our SPMM implementation in comparison to oneMKL's sparse::gemm~\cite{mklsparsegemm} implementation. Our implementation increases the performance on average by 1.37x and at most by 2.16x. These performance increases may be attributed to the high specialization of our code to the specific sizes and sparsities relevant in machine learning applications. The average relative performance gain compared to~\cite{Trevor2020} is approximately 2.4x with a maxium performance increase of approximately 4x (achieved on the 32k/8k/128/90\% case).
\include{figs/spmm_benchmark}

Figures~\ref{fig:fused_benchmark32} and~\ref{fig:fused_benchmark128} compares the performance of our FusedMM operation to oneMKL's dense::gemm and sparse::gemm operation. For the oneMKL case, we are not including the time required for the sampling after the dense matrix multiplication for the SDDMM operation. Our implementation increases the performance of MKL by approximately 2x on average and 3.25x at most. The fused operation achieves on average 1.3x the throughput of our SDDMM implementation and 1.074x of our SPMM implementation. The average performance gain is thus higher than in either of the cases before, underlining the importance of operator fusion for these kind of operations. The peak performance of 12,647 Gflops/s for this operations represents 48\% of the theoretical peak performance of the device.
\include{figs/fused_benchmark}

\subsection{Performance Analysis}
In this section, we investigate the efficiency of our implementation using the roofline model~\cite{advisorgpuroofline}. In particular, we examine the utilization of caches and how close we are to the theoretical peak of the underlying hardware. We selected a high-performant case with $M=8192$, $K=8192$, $N=128$, and $\alpha = 0.7$. 

\begin{figure}
    \centering
    \includegraphics[scale=0.5]{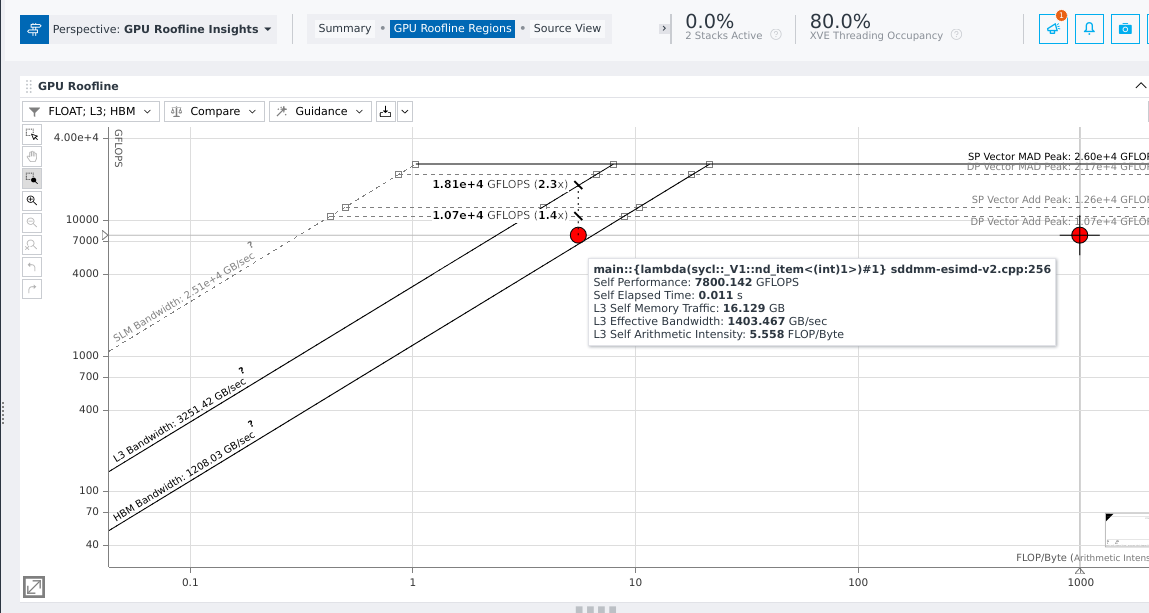}
    \caption{Roofline model illustrating performance of the SDDMM kernel for $M=8192$, $K=8192$, $N=128$, and $\alpha = 0.7$ } 
    \label{fig:sddmm_roofline}
\end{figure}

For this problem size, we observed a performance of $9,732$ Gflops/s for the SDDMM operation (cf. Figure~\ref{fig:sddmm_benchmark128}). The main computation in the SDDMM implementation is on line~15-16 of Figure~\ref{fig:sddmm-code-2}. Line~15 is a vector multiplication followed by a reduction. Hence, the performance of SDDMM is bounded by the "SP Vector Add Peak" of $12,600$ Gflops/s as shown in the roofline model in Figure~\ref{fig:sddmm_roofline}. The SDDMM performance of $9,732$ Gflops/s is 77\% of the theoretical peak. As indicated earlier, we ran a kernel in a loop over 20 iterations for all our experiments and selected the minimum execution time for plotting. Since the matrices are sufficiently small, they are cached in $L3$ after the first iteration. The minimum execution time thus corresponds to the iteration when most of the data is in L3. This is confirmed with the roofline model plotted by the Intel Advisor tool shown Figure~\ref{fig:sddmm_roofline}. The red dot on the right in Figure~\ref{fig:sddmm_roofline} represents the HBM traffic and it shows an amount of traffic to HBM which corresponds to only the first iteration. Note that the numbers shown in Figures~\ref{fig:sddmm_benchmark32} and ~\ref{fig:sddmm_benchmark128} are consistent since we used the same methodology to generate the MKL data: multiple consecutive runs in a loop and selected the minimum execution time. In separate experiments, which are not shown here, we observed a slowdown of close to $10\%$ for the first iteration compared to the rest of the iterations due to the necessary memory access to HBM instead of L3 cache. The aggregated performance over 20 iterations shown by the Advisor tool is $7800$ Gflops/s, which accounts for 10\% slow down for the first iteration and the overhead of the profiler. For the L3 we observed an arithmetic intensity of 5.6, which is significantly better compared to the theoretical {\em worst case} of the arithmetic intensity of 0.49. For the calculation of the worst case arithmetic intensity, we assume we load the columns of the $\mathbf{B}^T$ matrix always from L3; that is, there is no caching of the column in L1. In other words, the total number of bytes loaded for $\mathbf{B}^T$ from L3 is  $N \times nnz \times 4$ bytes. On the other hand, the theoretical peak arithmetic intensity, which would be possible under the assumption of infinite caches, is close to $57$. This number is calculated based on the assumption that the required data for the SDDMM operation is only loaded a single from the L3.

\begin{figure}
    \centering
    \includegraphics[scale=0.5]{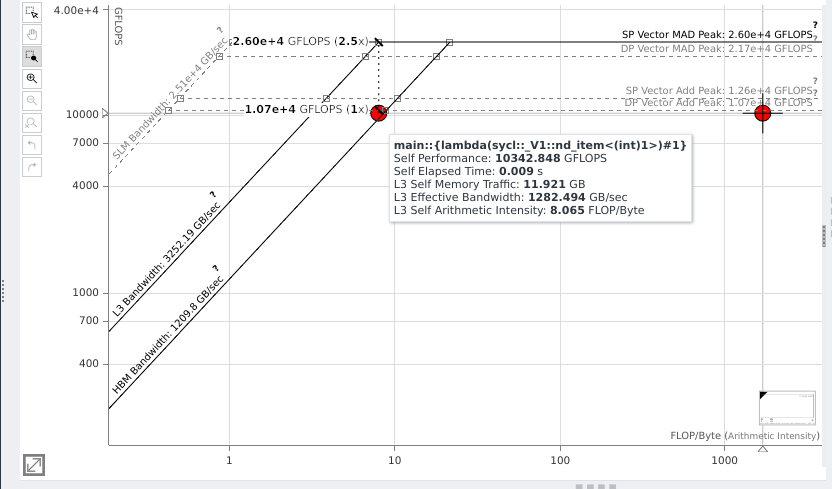}
    \caption{Roofline model illustrating performance of the SPMM kernel for $M=8192$, $K=8192$, $N=128$, and $\alpha = 0.7$ } 
    \label{fig:spmm_roofline}
\end{figure}

For our SPMM implementation, we observed a performance of $11,924$ Gflops/s for the high-performant case with $M=8192$, $K=8192$, $N=128$, and $\alpha = 0.7$. The memory access pattern and operation count for SPMM are similar to the SDDMM operation; however, we observed a better absolute performance for SPMM ($10,884$ Gflops/s) compared to SDDMM ($9,732$ Gflops/s). Thhe main reason for the increased performance compared to SDDMM is that the main computation in the SPMM implementation on line~25 is a multiply-add (MAD) operation; see Figure~\ref{fig:spmm-code-2}. While the absolute performance of SPMM is better, it achieves a lower relative performance compared to the theoretical peak performance of the hardware. This is due to the performance of SPMM being bounded by the "SP Vector MAD Peak" of $26,000$ Gflops/s as shown in the roofline model in Figure~\ref{fig:spmm_roofline} (We note that 2x the single precision vector add peak would actually amount to $25,200$ Gflops/s. We use $26,000$ Gflops/s to stay consistent with the Advisor tool). The SPMM performance is thus 37\% of the theoretical peak. Based on this data, we believe there is still room for improvement of the performance of this operation. As in the case of SDDMM, the matrices are cached in L3 after the first iteration. The aggregated performance over 20 iterations shown by the advisor tool is $10,342$ Gflops/s, which accounts for 10\% slow down for the first iteration and the overhead of the profiler. For the L3 cache we observed an arithmetic intensity of 8.06, which is better than in the SDDMM case and within the expectation of 0.49 (worst case) to $57$ (best case).

\section{Conclusion and Future Work}
In this contribution, we showed that highly optimized implementations of sparse matrix operations occurring in machine learning applications outperform existing approaches. In particular, we showed that the difference between the general SPMM implementation available in Intel's oneMKL can be outperformed by our implementation, which is highly specialized for relatively dense sparse matrices by up to 2.16x. Further, we showed a gap in Intel's oneMKL functionalities by not providing a dedicated SDDMM implementation. While the dense matrix multiply provided by MKL performs well on Intel's Max GPU, it suffers from the sparsity inherent in the SDDMM operation. When fusing SPMM and SDDMM operations, the performance gains increase further to up to a factor of 3.25. It is thus highly recommended to implement fused operations wherever possible in the context of machine learning applications.
We further showed that our sparse matrix operations can be run up to 10x faster on an Intel Data Center GPU compared to an optimized CUDA code on a NVIDIA V100 GPU. For our future work, we plan to explore sparse matrix operations utilizing bfloat16 data types and evaluate performance on a larger sparse matrix data set for different machine learning applications. 

\section*{Acknowledgements}
This effort has been supported by the Intel oneAPI Center of Excellence at Old Dominion University. 
We want to thank Xiao Zhu of Intel, who provided support throughout this project, making resources available whenever we needed them.

\bibliographystyle{unsrt}
\bibliography{spdl}

\end{document}

%% file: figs/sddmm_benchmark.tex
\begin{figure}
\centering
\pgfplotsset{scaled y ticks=false}
\begin{tikzpicture}
\begin{axis}[
    width=\textwidth,
    height=0.5\textwidth,
    ybar=0.5pt,
    bar width = 3pt,
    title={SDDMM Benchmark for $N=32$},
    ylabel={Gflops/s},
    ymin=0, 
    xtick=data,
    xticklabels={{1k,1k,32,70\%},{1k,1k,32,80\%},{1k,1k,32,90\%},{3k,1k,32,70\%},{3k,1k,32,80\%},{3k,1k,32,90\%},{4k,1k,32,70\%},{4k,1k,32,80\%},{4k,1k,32,90\%},{2k,2k,32,70\%},{2k,2k,32,80\%},{2k,2k,32,90\%},{6k,2k,32,70\%},{6k,2k,32,80\%},{6k,2k,32,90\%},{8k,2k,32,70\%},{8k,2k,32,80\%},{8k,2k,32,90\%},{4k,4k,32,70\%},{4k,4k,32,80\%},{4k,4k,32,90\%},{12k,4k,32,70\%},{12k,4k,32,80\%},{12k,4k,32,90\%},{16k,4k,32,70\%},{16k,4k,32,80\%},{16k,4k,32,90\%},{8k,8k,32,70\%},{8k,8k,32,80\%},{8k,8k,32,90\%},{24k,8k,32,70\%},{24k,8k,32,80\%},{24k,8k,32,90\%},{32k,8k,32,70\%},{32k,8k,32,80\%},{32k,8k,32,90\%}},
    xticklabel style={font=\small,rotate=90,xshift=0.5ex,yshift=0ex},
    legend pos=north west,
    ymajorgrids=true,
    yminorgrids=true,
    xmajorgrids=true,
    grid style=dashed,
    enlarge y limits={value=0.15, upper},
    enlarge x limits={abs=0.2cm},
]

\addplot[
    color=blue,
    fill=blue
    ]
    coordinates {
    (0,1996)
(1,1331)
(2,598)
(3,2691)
(4,1965)
(5,985)
(6,3216)
(7,2752)
(8,1779)
(9,3415)
(10,2922)
(11,1970)
(12,4048)
(13,3576)
(14,2883)
(15,4462)
(16,4114)
(17,3441)
(18,4472)
(19,4179)
(20,3552)
(21,4771)
(22,4549)
(23,4011)
(24,4797)
(25,4602)
(26,4082)
(27,4877)
(28,4729)
(29,4212)
(30,4375)
(31,4505)
(32,4315)
(33,4357)
(34,4290)
(35,4259)
    };
\addplot[
    color=red,
    fill=red
    ]
    coordinates {
    (0,1052)
(1,794)
(2,403)
(3,2012)
(4,1402)
(5,704)
(6,2637)
(7,1736)
(8,886)
(9,2669)
(10,1843)
(11,917)
(12,3941)
(13,2601)
(14,1317)
(15,4271)
(16,2791)
(17,1426)
(18,4106)
(19,2788)
(20,1382)
(21,4877)
(22,3240)
(23,1625)
(24,4544)
(25,3037)
(26,1492)
(27,4466)
(28,2978)
(29,1478)
(30,4553)
(31,3025)
(32,1502)
(33,4580)
(34,3020)
(35,1528)

    };
    \legend{Our SDDMM, MKL GEMM}
\end{axis}
\end{tikzpicture}
\caption{Comparison of our SDDMM method with MKL blas::gemm. The x-label indicates M/K/N/Sparsity. }
\label{fig:sddmm_benchmark32}
\end{figure}

\begin{figure}
\centering
\pgfplotsset{scaled y ticks=false}
\begin{tikzpicture}
\begin{axis}[
    width=\textwidth,
    height=0.5\textwidth,
    ybar=0.5pt,
    bar width = 3pt,
    title={SDDMM Benchmark for $N=128$},
    ylabel={Gflops/s},
    ymin=0,
    xtick=data,
    xticklabels={{1k,1k,128,70\%},{1k,1k,128,80\%},{1k,1k,128,90\%},{3k,1k,128,70\%},{3k,1k,128,80\%},{3k,1k,128,90\%},{4k,1k,128,70\%},{4k,1k,128,80\%},{4k,1k,128,90\%},{2k,2k,128,70\%},{2k,2k,128,80\%},{2k,2k,128,90\%},{6k,2k,128,70\%},{6k,2k,128,80\%},{6k,2k,128,90\%},{8k,2k,128,70\%},{8k,2k,128,80\%},{8k,2k,128,90\%},{4k,4k,128,70\%},{4k,4k,128,80\%},{4k,4k,128,90\%},{12k,4k,128,70\%},{12k,4k,128,80\%},{12k,4k,128,90\%},{16k,4k,128,70\%},{16k,4k,128,80\%},{16k,4k,128,90\%},{8k,8k,128,70\%},{8k,8k,128,80\%},{8k,8k,128,90\%},{24k,8k,128,70\%},{24k,8k,128,80\%},{24k,8k,128,90\%},{32k,8k,128,70\%},{32k,8k,128,80\%},{32k,8k,128,90\%}},
    xticklabel style={font=\small,rotate=90,xshift=0.5ex,yshift=0ex},
    legend pos=north west,
    ymajorgrids=true,
    yminorgrids=true,
    xmajorgrids=true,
    grid style=dashed,
    enlarge y limits={value=0.15, upper},
    enlarge x limits={abs=0.2cm},
]

\addplot[
    color=blue,
    fill=blue
    ]
    coordinates {
    (0,4553)
(1,3554)
(2,2142)
(3,6238)
(4,5246)
(5,3325)
(6,7954)
(7,7072)
(8,5191)
(9,6841)
(10,5640)
(11,3939)
(12,7710)
(13,6149)
(14,4430)
(15,9313)
(16,8621)
(17,7056)
(18,9411)
(19,8733)
(20,6685)
(21,9624)
(22,9185)
(23,7574)
(24,9659)
(25,9011)
(26,7587)
(27,9732)
(28,9374)
(29,7746)
(30,9389)
(31,8707)
(32,7837)
(33,9153)
(34,8565)
(35,7611)

    };
\addplot[
    color=red,
    fill=red
    ]
    coordinates {
    (0,3206)
(1,2151)
(2,1083)
(3,4663)
(4,3121)
(5,1557)
(6,5092)
(7,3416)
(8,1717)
(9,5133)
(10,3441)
(11,1710)
(12,6311)
(13,4256)
(14,2123)
(15,6531)
(16,4357)
(17,2185)
(18,6509)
(19,4346)
(20,2164)
(21,6888)
(22,4583)
(23,2291)
(24,6922)
(25,4640)
(26,2319)
(27,6862)
(28,4577)
(29,2283)
(30,7112)
(31,4724)
(32,2346)
(33,7180)
(34,4835)
(35,2423)

    };
    \legend{Our SDDMM, MKL GEMM}
\end{axis}
\end{tikzpicture}
\caption{Comparison of our SDDMM method with MKL blas::gemm. The x-label indicates M/K/N/Sparsity. }
\label{fig:sddmm_benchmark128}
\end{figure}

%% file: figs/spmm_benchmark.tex
\begin{figure}
\centering
\pgfplotsset{scaled y ticks=false}
\begin{tikzpicture}
\begin{axis}[
    width=\textwidth,
    height=0.5\textwidth,
    ybar=0.5pt,
    bar width = 3pt,
    title={SpMM Benchmark for $N=32$},
    ylabel={Gflops/s},
    ymin=0, 
    xtick=data,
    xticklabels={{1k,1k,32,70\%},{1k,1k,32,80\%},{1k,1k,32,90\%},{3k,1k,32,70\%},{3k,1k,32,80\%},{3k,1k,32,90\%},{4k,1k,32,70\%},{4k,1k,32,80\%},{4k,1k,32,90\%},{2k,2k,32,70\%},{2k,2k,32,80\%},{2k,2k,32,90\%},{6k,2k,32,70\%},{6k,2k,32,80\%},{6k,2k,32,90\%},{8k,2k,32,70\%},{8k,2k,32,80\%},{8k,2k,32,90\%},{4k,4k,32,70\%},{4k,4k,32,80\%},{4k,4k,32,90\%},{12k,4k,32,70\%},{12k,4k,32,80\%},{12k,4k,32,90\%},{16k,4k,32,70\%},{16k,4k,32,80\%},{16k,4k,32,90\%},{8k,8k,32,70\%},{8k,8k,32,80\%},{8k,8k,32,90\%},{24k,8k,32,70\%},{24k,8k,32,80\%},{24k,8k,32,90\%},{32k,8k,32,70\%},{32k,8k,32,80\%},{32k,8k,32,90\%}},
    xticklabel style={font=\small,rotate=90,xshift=0.5ex,yshift=0ex},
    legend pos=north west,
    ymajorgrids=true,
    yminorgrids=true,
    xmajorgrids=true,
    grid style=dashed,
    enlarge y limits={value=0.15, upper},
    enlarge x limits={abs=0.2cm},
]

\addplot[
    color=blue,
    fill=blue
    ]
    coordinates {
    (0,1120)
(1,938)
(2,604)
(3,3844)
(4,3220)
(5,2127)
(6,4973)
(7,4239)
(8,2888)
(9,2922)
(10,2448)
(11,1749)
(12,4897)
(13,4857)
(14,3868)
(15,6206)
(16,6289)
(17,4887)
(18,6346)
(19,5777)
(20,4499)
(21,6603)
(22,6095)
(23,4964)
(24,6585)
(25,6166)
(26,5038)
(27,6730)
(28,6313)
(29,5197)
(30,5819)
(31,5763)
(32,5351)
(33,5740)
(34,5540)
(35,5171)

    };
\addplot[
    color=red,
    fill=red
    ]
    coordinates {
    (0,696)
(1,605)
(2,376)
(3,2221)
(4,1689)
(5,985)
(6,2983)
(7,2244)
(8,1357)
(9,2073)
(10,1722)
(11,1116)
(12,3635)
(13,3115)
(14,2163)
(15,4750)
(16,4052)
(17,2856)
(18,4935)
(19,4362)
(20,3056)
(21,5605)
(22,5083)
(23,3942)
(24,5674)
(25,5211)
(26,4093)
(27,5906)
(28,5435)
(29,4402)
(30,5079)
(31,5061)
(32,4733)
(33,4971)
(34,4815)
(35,4591)

    };
    \legend{Our SPMM, MKL SPMM}
\end{axis}
\end{tikzpicture}
\caption{Comparison of our SPMM method with MKL sparse::gemm. The x-label indicates M/K/N/Sparsity. }
\label{fig:spmm_benchmark32}
\end{figure}

\begin{figure}
\centering
\pgfplotsset{scaled y ticks=false}
\begin{tikzpicture}
\begin{axis}[
    width=\textwidth,
    height=0.5\textwidth,
    ybar=0.5pt,
    bar width = 3pt,
    title={SpMM Benchmark for $N=128$},
    ylabel={Gflops/s},
    ymin=0,
    xtick=data,
    xticklabels={{1k,1k,128,70\%},{1k,1k,128,80\%},{1k,1k,128,90\%},{3k,1k,128,70\%},{3k,1k,128,80\%},{3k,1k,128,90\%},{4k,1k,128,70\%},{4k,1k,128,80\%},{4k,1k,128,90\%},{2k,2k,128,70\%},{2k,2k,128,80\%},{2k,2k,128,90\%},{6k,2k,128,70\%},{6k,2k,128,80\%},{6k,2k,128,90\%},{8k,2k,128,70\%},{8k,2k,128,80\%},{8k,2k,128,90\%},{4k,4k,128,70\%},{4k,4k,128,80\%},{4k,4k,128,90\%},{12k,4k,128,70\%},{12k,4k,128,80\%},{12k,4k,128,90\%},{16k,4k,128,70\%},{16k,4k,128,80\%},{16k,4k,128,90\%},{8k,8k,128,70\%},{8k,8k,128,80\%},{8k,8k,128,90\%},{24k,8k,128,70\%},{24k,8k,128,80\%},{24k,8k,128,90\%},{32k,8k,128,70\%},{32k,8k,128,80\%},{32k,8k,128,90\%}},
    xticklabel style={font=\small,rotate=90,xshift=0.5ex,yshift=0ex},
    legend pos=north west,
    ymajorgrids=true,
    yminorgrids=true,
    xmajorgrids=true,
    grid style=dashed,
    enlarge y limits={value=0.15, upper},
    enlarge x limits={abs=0.2cm},
]

\addplot[
    color=blue,
    fill=blue
    ]
    coordinates {
    (0,2935)
(1,2475)
(2,1774)
(3,7687)
(4,6742)
(5,4873)
(6,9896)
(7,8699)
(8,6262)
(9,6130)
(10,5452)
(11,4046)
(12,8287)
(13,7587)
(14,6132)
(15,10884)
(16,9813)
(17,7439)
(18,11083)
(19,10040)
(20,7772)
(21,11334)
(22,10294)
(23,8010)
(24,11334)
(25,10406)
(26,8110)
(27,11482)
(28,10506)
(29,8214)
(30,10977)
(31,10197)
(32,8346)
(33,10975)
(34,10135)
(35,8182)

    };
\addplot[
    color=red,
    fill=red
    ]
    coordinates {
    (0,2615)
(1,2102)
(2,1278)
(3,5060)
(4,4239)
(5,2671)
(6,6379)
(7,5268)
(8,3528)
(9,6605)
(10,5596)
(11,3765)
(12,7391)
(13,6352)
(14,5073)
(15,8214)
(16,7234)
(17,5268)
(18,8504)
(19,7625)
(20,5744)
(21,8955)
(22,8176)
(23,6366)
(24,8993)
(25,8247)
(26,6457)
(27,8897)
(28,8474)
(29,6766)
(30,8460)
(31,7811)
(32,6717)
(33,8406)
(34,7794)
(35,6560)

    };
    \legend{Our SPMM, MKL SPMM}
\end{axis}
\end{tikzpicture}
\caption{Comparison of our SPMM method with MKL sparse::gemm. The x-label indicates M/K/N/Sparsity. }
\label{fig:spmm_benchmark128}
\end{figure}

%% file: figs/fused_benchmark.tex
\begin{figure}
\centering
\pgfplotsset{scaled y ticks=false}
\begin{tikzpicture}
\begin{axis}[
    width=\textwidth,
    height=0.5\textwidth,
    ybar=0.5pt,
    bar width = 3pt,
    title={FusedMM Benchmark for $N=32$},
    ylabel={Gflops/s},
    ymin=0, 
    xtick=data,
    xticklabels={{1k,1k,32,70\%},{1k,1k,32,80\%},{1k,1k,32,90\%},{3k,1k,32,70\%},{3k,1k,32,80\%},{3k,1k,32,90\%},{4k,1k,32,70\%},{4k,1k,32,80\%},{4k,1k,32,90\%},{2k,2k,32,70\%},{2k,2k,32,80\%},{2k,2k,32,90\%},{6k,2k,32,70\%},{6k,2k,32,80\%},{6k,2k,32,90\%},{8k,2k,32,70\%},{8k,2k,32,80\%},{8k,2k,32,90\%},{4k,4k,32,70\%},{4k,4k,32,80\%},{4k,4k,32,90\%},{12k,4k,32,70\%},{12k,4k,32,80\%},{12k,4k,32,90\%},{16k,4k,32,70\%},{16k,4k,32,80\%},{16k,4k,32,90\%},{8k,8k,32,70\%},{8k,8k,32,80\%},{8k,8k,32,90\%},{24k,8k,32,70\%},{24k,8k,32,80\%},{24k,8k,32,90\%},{32k,8k,32,70\%},{32k,8k,32,80\%},{32k,8k,32,90\%}},
    xticklabel style={font=\small,rotate=90,xshift=0.5ex,yshift=0ex},
    legend pos=north west,
    ymajorgrids=true,
    yminorgrids=true,
    xmajorgrids=true,
    grid style=dashed,
    enlarge y limits={value=0.15, upper},
    enlarge x limits={abs=0.2cm},
]

\addplot[
    color=blue,
    fill=blue
    ]
    coordinates {
(0,1369)
(1,1093)
(2,686)
(3,4300)
(4,3692)
(5,2495)
(6,5115)
(7,4432)
(8,2944)
(9,3629)
(10,3094)
(11,2095)
(12,5544)
(13,4831)
(14,3930)
(15,6955)
(16,6334)
(17,5058)
(18,7079)
(19,6522)
(20,5220)
(21,7549)
(22,7079)
(23,5811)
(24,7577)
(25,7169)
(26,5899)
(27,7691)
(28,7283)
(29,6196)
(30,7625)
(31,7418)
(32,6418)
(33,7609)
(34,7286)
(35,6429)
    };
\addplot[
    color=red,
    fill=red
    ]
    coordinates {
(0,777)
(1,592)
(2,314)
(3,1960)
(4,1402)
(5,767)
(6,2520)
(7,1822)
(8,980)
(9,2151)
(10,1605)
(11,916)
(12,3644)
(13,2704)
(14,1571)
(15,4310)
(16,3204)
(17,1795)
(18,4422)
(19,3291)
(20,1867)
(21,4211)
(22,3406)
(23,2140)
(24,3829)
(25,3327)
(26,2125)
(27,3962)
(28,3293)
(29,2116)
(30,4528)
(31,3503)
(32,2189)
(33,4613)
(34,3613)
(35,2212)
    };
    \legend{Our Fused, MKL SDDMM+SPMM}
\end{axis}
\end{tikzpicture}
\caption{Comparison of our FusedMM method with MKL. The x-label indicates M/K/N/Sparsity. }
\label{fig:fused_benchmark32}
\end{figure}

\begin{figure}
\centering
\pgfplotsset{scaled y ticks=false}
\begin{tikzpicture}
\begin{axis}[
    width=\textwidth,
    height=0.5\textwidth,
    ybar=0.5pt,
    bar width = 3pt,
    title={FusedMM Benchmark for $N=128$},
    ylabel={Gflops/s},
    ymin=0,
    xtick=data,
    xticklabels={{1k,1k,128,70\%},{1k,1k,128,80\%},{1k,1k,128,90\%},{3k,1k,128,70\%},{3k,1k,128,80\%},{3k,1k,128,90\%},{4k,1k,128,70\%},{4k,1k,128,80\%},{4k,1k,128,90\%},{2k,2k,128,70\%},{2k,2k,128,80\%},{2k,2k,128,90\%},{6k,2k,128,70\%},{6k,2k,128,80\%},{6k,2k,128,90\%},{8k,2k,128,70\%},{8k,2k,128,80\%},{8k,2k,128,90\%},{4k,4k,128,70\%},{4k,4k,128,80\%},{4k,4k,128,90\%},{12k,4k,128,70\%},{12k,4k,128,80\%},{12k,4k,128,90\%},{16k,4k,128,70\%},{16k,4k,128,80\%},{16k,4k,128,90\%},{8k,8k,128,70\%},{8k,8k,128,80\%},{8k,8k,128,90\%},{24k,8k,128,70\%},{24k,8k,128,80\%},{24k,8k,128,90\%},{32k,8k,128,70\%},{32k,8k,128,80\%},{32k,8k,128,90\%}},
    xticklabel style={font=\small,rotate=90,xshift=0.5ex,yshift=0ex},
    legend pos=north west,
    ymajorgrids=true,
    yminorgrids=true,
    xmajorgrids=true,
    grid style=dashed,
    enlarge y limits={value=0.15, upper},
    enlarge x limits={abs=0.2cm},
]

\addplot[
    color=blue,
    fill=blue
    ]
    coordinates {
(0,2793)
(1,2306)
(2,1629)
(3,8503)
(4,7373)
(5,5255)
(6,10344)
(7,9130)
(8,6530)
(9,7094)
(10,6084)
(11,4324)
(12,7482)
(13,6927)
(14,5602)
(15,11793)
(16,10598)
(17,7136)
(18,11971)
(19,10948)
(20,6353)
(21,12443)
(22,11369)
(23,6572)
(24,12130)
(25,11269)
(26,6541)
(27,12647)
(28,11579)
(29,4688)
(30,11936)
(31,11173)
(32,5098)
(33,11939)
(34,10987)
(35,4991)
    };
\addplot[
    color=red,
    fill=red
    ]
    coordinates {
(0,2525)
(1,1911)
(2,1035)
(3,4498)
(4,3296)
(5,1850)
(6,5337)
(7,3856)
(8,2098)
(9,5396)
(10,3939)
(11,2152)
(12,6373)
(13,4820)
(14,2817)
(15,6986)
(16,5215)
(17,2933)
(18,7054)
(19,5266)
(20,2976)
(21,7433)
(22,5643)
(23,3233)
(24,7454)
(25,5681)
(26,3289)
(27,7479)
(28,5692)
(29,3302)
(30,7334)
(31,5711)
(32,3341)
(33,7320)
(34,5632)
(35,3302)
    };
    \legend{Our Fused, MKL SDDMM+SpMM}
\end{axis}
\end{tikzpicture}
\caption{Comparison of our FusedMM method with MKL. The x-label indicates M/K/N/Sparsity. }
\label{fig:fused_benchmark128}
\end{figure}